\documentclass[11pt]{article}

\usepackage[preprint]{acl}

\usepackage{times}
\usepackage{latexsym}

\usepackage[T1]{fontenc}

\usepackage[utf8]{inputenc}

\usepackage{microtype}

\usepackage{inconsolata}

\usepackage{graphicx}

\usepackage{amsmath} 
\usepackage{amssymb}  

\usepackage{csquotes} 
\usepackage{booktabs} 
\usepackage{multirow} 
\usepackage{stfloats} 

\usepackage{graphicx}
\usepackage{subcaption}  

\usepackage{url}            

\usepackage[linesnumbered,ruled,vlined]{algorithm2e}

\usepackage{xcolor}
\usepackage{listings}
\usepackage{courier} 
\usepackage[most]{tcolorbox} 
\usepackage{cuted}  

\definecolor{codegreen}{rgb}{0,0.6,0}       
\definecolor{codegray}{rgb}{0.5,0.5,0.5}    
\definecolor{codepurple}{rgb}{0.58,0,0.82}  
\definecolor{backcolour}{rgb}{0.95,0.95,0.92} 
\definecolor{keywordblue}{rgb}{0,0,0.6}     
\definecolor{stringred}{rgb}{0.8,0,0}       

\definecolor{promptframe}{rgb}{0.4, 0.6, 0.6} 
\definecolor{promptback}{rgb}{0.93, 0.95, 0.95} 

\lstdefinestyle{mystyle}{
    backgroundcolor=\color{backcolour},
    commentstyle=\color{codegreen}\itshape, 
    keywordstyle=\color{keywordblue}\bfseries, 
    numberstyle=\tiny\color{codegray},      
    stringstyle=\color{stringred},          
    basicstyle=\ttfamily\footnotesize,      
    breakatwhitespace=false,
    breaklines=true,                        
    captionpos=b,                           
    keepspaces=true,
    numbers=left,                           
    numbersep=5pt,
    showspaces=false,
    showstringspaces=false,
    showtabs=false,
    tabsize=4,
    frame=single,                           
    rulecolor=\color{black}                 
}

\newtcolorbox{promptboxcross}[1][]{
    enhanced,
    breakable,
    width=\linewidth,
    colback=promptback,
    colframe=promptframe,
    fontupper=\ttfamily\small,
    attach boxed title to top left={
        xshift=5mm, 
        yshift=-3mm,
        yshifttext=-1mm
    },
    boxed title style={
        colback=promptback,
        colframe=promptframe!50,
        sharp corners,
    },
    coltitle=black,               
    title={\texttt{>\_} Prompt},  
    sharp corners=southwest,
    arc=3mm,
    before upper={\obeylines},
    #1
}

\newenvironment{promptbox*}[1][]{%
    \begin{strip}
    \begin{promptboxcross}[#1]
}{%
    \end{promptboxcross}
    \end{strip}
}

\title{BiFE: Search-Efficient Discovery of CPU-Only Branching Policies via LLM-based Bi-Fidelity Evolution}

\author{
  \textbf{Ce Zhang\textsuperscript{1,2}},
  \textbf{Bin Zhang\textsuperscript{1}},
  \textbf{Zhiwei Xu\textsuperscript{3}},
  \textbf{Hao Chen\textsuperscript{1,2}},
  \textbf{Xinyue Lu\textsuperscript{1,2}},\\
  \textbf{Shanwei Fan\textsuperscript{1,2}},
  \textbf{Yingxuan Teng\textsuperscript{1,2}},
  \textbf{Guoliang Fan\textsuperscript{1}}
\\
  \textsuperscript{1}Institute of Automation, Chinese Academy of Sciences \\
  \textsuperscript{2}School of Artificial Intelligence, University of Chinese Academy of Sciences \\
  \textsuperscript{3}Shandong University\\
    \texttt{\{zhangce2023, zhangbin2020, chenhao2022, luxinyue2023,} \\
  \texttt{fanshanwei2024, tengyingxuan2024, guoliang.fan\}@ia.ac.cn} \\
  \texttt{zhiwei\_xu@sdu.edu.cn}
}

\begin{document}
\maketitle
\begin{abstract}
In branch-and-bound (B\&B) for mixed-integer linear programming (MILP), branching variable selection critically impacts efficiency. Existing neural branching policies often require GPU inference, while CPU-efficient symbolic expressions lack the representational capacity for complex logic.
Large Language Model (LLM)-generated code provides a flexible search space for designing lightweight branching rules with diverse algorithmic logic.
To discover effective rules within LLM-based evolutionary frameworks, a core challenge arises: full B\&B evaluation on real instances is prohibitively expensive, whereas offline imitation learning suffers from distribution shift. To address this, we introduce a Bi-Fidelity Evolutionary framework (BiFE). It employs low-fidelity imitation scores as a rapid pre-screener and selectively applies high-fidelity on-instance evaluation only to elite candidates, effectively balancing search efficiency with performance reliability. Experiments validate both the search efficiency of BiFE and the competitiveness of its discovered rules, which outperform the SCIP solver and other baselines on CPUs, and even surpass certain GPU-based neural policies.
\end{abstract}

\section{Introduction}

Combinatorial optimization problems represent a class of NP-hard problems with broad applications in transportation~\cite{barnhart2006handbooks}, scheduling~\cite{chen2010integrated}, and planning~\cite{pochet2006production}.
Many such problems can be expressed as mixed-integer linear programs (MILPs) and solved exactly through the branch-and-bound (B\&B) framework~\cite{land2010automatic}.
B\&B proceeds by iteratively branching on variables that deviate from integrality constraints to yield subproblem nodes, meanwhile updating the global upper and lower bounds and pruning infeasible or suboptimal nodes, until it converges to the global optimal solution.
This method underlies high-performance MILP solvers, including SCIP~\cite{scip}, Gurobi~\cite{gurobi}, and COPT~\cite{copt}.

Within the B\&B framework, branching variable selection is one of the key determinants for MILP solving efficiency. Recently, many studies have developed learning-based branching policies to supplant handcrafted heuristics~\cite{gasse2019exact,zarpellon2021parameterizing,scavuzzo2022tmdp}.
Neural policies learn from instance distributions and can significantly reduce tree size, but they typically rely on GPUs for fast inference. Since real-world MILP solving often runs on CPU-only environments, Gupta et al.~\cite{gupta2020hybrid} propose a CPU-efficient scheme using a complex network only at the root and lightweight networks elsewhere. Later, Kuang et al.~\cite{kuang2024rethinking,kuang2024towards} further improve CPU efficiency by representing branching rules as simple symbolic expressions discovered through neuro-symbolic search, which also brings interpretability.

Symbolic operations, however, are inherently limited in representational capacity. Effective branching policies may demand more complex logic than a fixed expression grammar can capture. Large Language Model (LLM)-based algorithm-design frameworks~\cite{Paredes2024FunSearch,liu2024EoH,ye2024reevo} offer a natural way to search for richer lightweight rules through evolutionary code generation. LLM4Branch~\cite{hou2026llm4branch} applies this paradigm to branching by combining LLM-generated program skeletons with zeroth-order parameter optimization based on direct solver feedback.

Here, fitness evaluation poses a core challenge in the MILP setting. Directly testing each candidate heuristic on real MILP instances demands full B\&B solves, which becomes prohibitively expensive when searching a vast code space. A tempting alternative is to inherit the imitation learning paradigm from neural branching policies, mimicking strong-branching decisions on offline-collected data~\cite{gasse2019exact,gupta2020hybrid}.
Yet this suffers from distribution shift: the heuristic's own deployment trajectory can diverge significantly from the offline demonstrations, so high imitation accuracy offers no guarantee that the resulting rule is truly CPU-efficient and yields strong solving performance.

To bridge this gap, we propose a \textbf{Bi}-\textbf{F}idelity \textbf{E}volutionary framework inspired by multi-fidelity optimization~\cite{li2026mfo}, termed BiFE. The low-fidelity imitation score serves as a rapid pre-screening filter, while high-fidelity evaluation on real MILPs is selectively applied only to elite candidates. This design preserves the efficiency of surrogate-based search while retaining the reliability of on-instance testing, enabling practical evolution of branching policies in the LLM-generated code space.
Our contributions can be summarized as follows:
\begin{itemize}
\item Branching rule discovery is formulated as a bi-fidelity evaluation problem: strong-branching imitation is efficient but subject to distribution shift, whereas solver feedback is reliable but costly.
\item We propose BiFE, which introduces elite-prioritized admission and selection to achieve an effective balance between search efficiency and the performance of discovered branching rules
\item Experiments show that BiFE improves rule-discovery efficiency while producing branching rules that outperform comparable baselines and the SCIP solver on CPUs, and even surpass certain neural branching policies evaluated on GPUs.
\end{itemize}

\section{Related Work}
\noindent\textbf{Deep Learning for MILP.}
The neural network acceleration of MILP solving can be roughly divided into two directions~\citep{bengio2021machine,scavuzzo2024machine}: (1)~Replacing traditional heuristic rules with neural networks within the B\&B framework, such as variable selection~\citep{Khalil2016Learning,gasse2019exact,gupta2020hybrid,zarpellon2021parameterizing,gupta2022lookback}, node selection~\citep{he2014learning,labassi2022learning,zhang2025learning}, cut selection~\citep{tang2020reinforcement,huang2022learning,wang2023learning,arjun2026miracle} and backdoor prediction~\citep{ferber2022backdoor,cai2024backdoorcl}. (2)~Using neural networks as a primal heuristic to obtain a high-quality feasible solution as the initial primal bound. This category includes two approaches: solution prediction~\citep{ding2020accelerating,han2023gnn,huang2024contrastive,liu2025apollomilp,pu2025rome} and neighborhood selection~\citep{wu2021learning,sonnerat2022learning,huang2023search}.
We focus on CPU-efficient branching policies, a sub-direction of variable selection progressing from lightweight networks~\citep{gupta2020hybrid} to symbolic rules~\citep{kuang2024rethinking,kuang2024towards}. Our method introduces a new paradigm beyond both trends, possessing competitive representational power and near-symbolic computational efficiency.

\noindent\textbf{LLM for MILP.}
Early research on LLM-based algorithm design~\citep{Paredes2024FunSearch,liu2024EoH,ye2024reevo} primarily focuses on specific components of particular combinatorial optimization problems, such as constructive heuristics for the traveling salesman problem.
More recently, some studies have extended LLMs to heuristic design for general MILPs, concentrating on primal heuristics such as diving heuristics~\citep{zhou2024llmsolver,zhang2026dhevo} and neighborhood selection~\citep{ye2025llm4lns}, or cutting plane selection~\citep{li2026strcmp} and design~\citep{yazdani2025evocut}.
For variable selection, \citet{zheng2026llm4scheduling} uses LLMs to schedule handcrafted heuristic rules online. Closely contemporaneous work, LLM4Branch~\citep{hou2026llm4branch}, combines program-skeleton generation with zeroth-order parameter tuning based on direct solver feedback. BiFE instead focuses on evaluation allocation during program evolution, using imitation as a cheap screening fidelity and selectively applying costly solver evaluations to correct the surrogate.

\section{Preliminaries}
\label{sec:preliminaries}
\noindent\textbf{B\&B Algorithm.}
The standard form of MILPs is:
$\arg\min_{\mathbf{x}}\left\{\mathbf{c}^{\top}\mathbf{x}\mid\mathbf{A}\mathbf{x}\leq\mathbf{b},\mathbf{x}\in\mathbb{Z}^{p}\times\mathbb{R}^{n-p}\right\}$,
where the vector $\mathbf{x}$ represents $n$ variables to be optimized, with $p$ being the number of integer variables.
$\mathbf{A},\mathbf{b},\mathbf{c}$ represent  constraint matrix, constraint right term, and objective coefficient.
The B\&B algorithm for solving MILPs consists of three steps: branching, bounding, and pruning. Starting from the relaxed root node solution, this method iteratively branches on fractional variables $x_i = b_i$ by adding $x_i \leq \left[b_i\right]$ and $x_i \geq \left[b_i\right] + 1$, maintains global primal and dual bounds (i.e., upper and lower bounds), and prunes nodes that are suboptimal or infeasible. The procedure terminates when the upper and lower bounds align, yielding the optimal solution.

\noindent\textbf{Handcrafted Branching Rules.}
In the B\&B algorithm, variable selection involves choosing one variable from several fractional candidates. 
This decision significantly affects the total node count, which in turn influences overall solving time. Several handcrafted branching rules are discussed in~\citep{achterberg2005branching}. Pseudo-cost branching uses historical data to guide current decisions, but suffers from poor accuracy early in the search tree.
Strong branching evaluates the dual bound improvement from creating child nodes for all candidates. Although effective in reducing the node count, its high computational cost of solving all child nodes undermines its purpose of acceleration. However, this property makes strong branching suitable as an expert for imitation learning in neural branching policies~\citep{gasse2019exact,gupta2020hybrid}. In MILP solvers, a hybrid approach is typical: strong branching initializes the process, while pseudo-cost branching takes over later.

\section{Methodology}
\label{section:method}
In this section, we formally introduce BiFE, a bi-fidelity evolutionary framework in which a low-fidelity surrogate based on strong-branching imitation serves as an efficient filter, while high-fidelity MILP solving evaluation is selectively applied to elite candidates. Candidate heuristics are generated by an LLM, and an elite-prioritized evolutionary mechanism allocates the two fidelities, enabling the discovery of CPU-efficient branching rules. The overall framework is illustrated in Figure~\ref{fig:overall}.
The pseudocode of BiFE is provided in Appendix~\ref{appendix:pseudocode}.

\subsection{Bi-Fidelity Formulation}

We formulate branching rule discovery as a black-box optimization problem over a heuristic space $\mathcal{H}$.
Each candidate heuristic $\hat{f} \in \mathcal{H}$ assigns scores to candidate branching variables, and the variable with the highest score is selected during B\&B search.
The objective is to identify a branching heuristic that maximizes the solving performance over a distribution of MILP instances:
\begin{equation}
\hat{f}^{\star} = \arg\max_{\hat{f}\in\mathcal{H}} F_{HF}(\hat{f}),
\end{equation}
where the high-fidelity objective $F_{HF}$ is defined as
\begin{equation}
F_{HF}(\hat{f})
=
-\frac{1}{|\mathcal{M}|}
\sum_{m\in\mathcal{M}}
\mathcal{J}(m,\hat{f}),
\end{equation}
with $\mathcal{M}$ denoting a set of MILP instances and $\mathcal{J}(m,\hat{f})$ representing the solving cost of heuristic $\hat{f}$ on instance $m$, measured by metrics such as solving time or node count.

However, directly optimizing $F_{HF}$ is computationally prohibitive.
Each evaluation requires deploying the heuristic inside a solver and executing complete B\&B procedures on multiple MILP instances, and the sheer number of candidates evaluated during evolutionary search renders relying solely on high-fidelity evaluations impractical.
To address this, we introduce a lightweight low-fidelity objective $F_{LF}: \mathcal{H} \rightarrow \mathbb{R}$.
By design, $F_{LF}$ is substantially cheaper to evaluate than $F_{HF}$, yet it cannot perfectly reproduce the scoring of $F_{HF}$ on candidate heuristics, as a perfect surrogate is typically unattainable in practice.
This motivates a two-tier evaluation strategy: the low-fidelity objective acts as an efficient filter, rapidly discarding unpromising candidates, while the expensive $F_{HF}$ is reserved as a fine-grained verifier applied only to the surviving elite, correcting any inaccuracies introduced by the surrogate approximation.

\subsection{Low-Fidelity Surrogate}

To construct the low-fidelity objective, we define heuristic quality according to the consistency between candidate branching decisions and strong-branching demonstrations.
As mentioned in Preliminaries, strong branching yields accurate results but is time-consuming, making it unsuitable for direct online use. Nevertheless, it is well-suited for offline collection as expert experience~\cite{gasse2019exact,gupta2020hybrid}.

During data collection, strong branching is used as the branching rule to solve a class of instances, and data pairs $(\mathbf{X}_t,a^{\star}_t)$ are collected throughout the process. Here $\mathbf{X}_t\in \mathbb{R}^{|\mathcal{C}|\times 72}$ denotes the features of candidate variables at the current branching step $t$, $\mathcal{C}$ denotes the set of candidate variables, and $|\mathcal{C}|$ is the number of candidate variables. Specifically, we adopt the 72-dimensional features designed by~\cite{Khalil2016Learning}, whose 18 static and 54 dynamic dimensions are clearly defined to encapsulate critical guidance for branching variable selection.
$a^{\star}=\arg\max_{c\in \mathcal{C}}\mathbf{s}_{sb}^c$ is the branching variable selected by strong branching, $\mathbf{s}_{sb}\in \mathbb{R}^{|\mathcal{C}|}$ denotes the strong-branching scores.
Let $\mathcal{D} = \{ (\mathbf{X}_t,a^{\star}_t) \}_{t=1}^{T}$ denote the set of $T$ data pairs, the low-fidelity surrogate is defined as follows:
\begin{equation}
    \label{equation:F_LF}
F_{LF}(\hat{f}) = \frac{1}{|\mathcal{D}|} \sum_{(\mathbf{X},a^\star)\in \mathcal{D}} \mathbb{I}(\arg\max_{c\in\mathcal{C}} \hat{f}(\mathbf{X}^c) = a^\star),
\end{equation}
where $\mathbb{I}(\cdot)$ is the indicator function, returning 1 if its argument is true and 0 otherwise, and $\hat{f}: \mathbb{R}^{72} \rightarrow \mathbb{R}^1$. This formulation reflects the same principle behind the cross-entropy loss used in imitation learning for neural branching~\cite{gasse2019exact}: the goal is to match the expert's decisions rather than regress its exact scores. Accordingly, we focus solely on the optimal action $a^\star$ and disregard all other scores. Computing $F_{LF}$ requires only evaluation on $\mathcal{D}$, so its cost is negligible.

The inability of $F_{LF}$ to perfectly reproduce the scoring of $F_{HF}$ stems from two inherent factors:
(1) Strong branching itself is myopic, optimizing only the immediate dual bound improvement at each step, so even perfect imitation does not guarantee globally optimal decisions.
(2) $\mathcal{D}$ is collected from trajectories of strong branching, while a learned heuristic generates its own trajectory during deployment. This distribution shift induces a mismatch between $F_{LF}$ and $F_{HF}$: high $F_{LF}$ on $\mathcal{D}$ does not guarantee strong solving performance.

\begin{figure*}[ht]
    \centering
    \includegraphics[width=\linewidth]{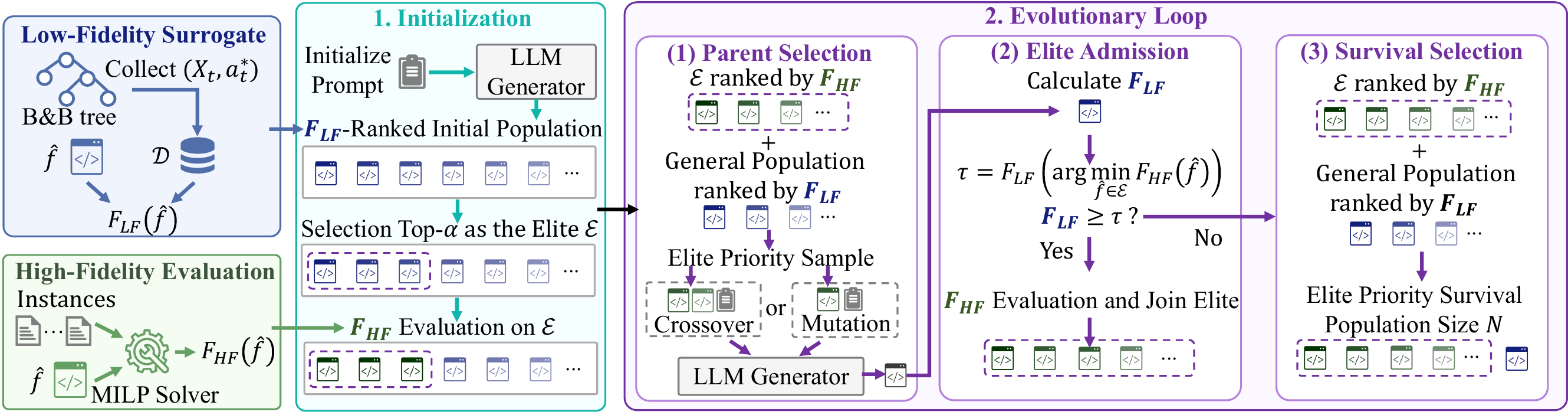}
    \caption{The overall framework of BiFE. To balance search efficiency and rule quality, $F_{HF}$ evaluates heuristics via expensive MILP solving while $F_{LF}$ offers a cheap surrogate by measuring consistency with strong-branching decisions.
    An LLM generator produces new heuristics through three operations: initialization, crossover, and mutation.
    The evolutionary process has two stages: (1) Initialization: the LLM generates an initial population, ranked by $F_{LF}$ with the top $\alpha$ forming the elite set $\mathcal{E}$ for $F_{HF}$ evaluation. (2) Evolutionary Loop: parent selection is elite-prioritized, where elites are ranked by $F_{HF}$ and others by $F_{LF}$, and selected parents are passed to the LLM for crossover or mutation. New candidates with $F_{LF}$ exceeding threshold $\tau$ undergo $F_{HF}$ evaluation and are admitted into $\mathcal{E}$. Survival selection retains all elites and fills the remaining slots from the general population.
    This selective allocation lets BiFE explore broadly with $F_{LF}$ while using $F_{HF}$ to correct the surrogate on promising candidates.
    }
    \label{fig:overall}
    \vspace{-4pt}
\end{figure*}

\begin{figure}[t]
    \centering
    \includegraphics[width=\linewidth]{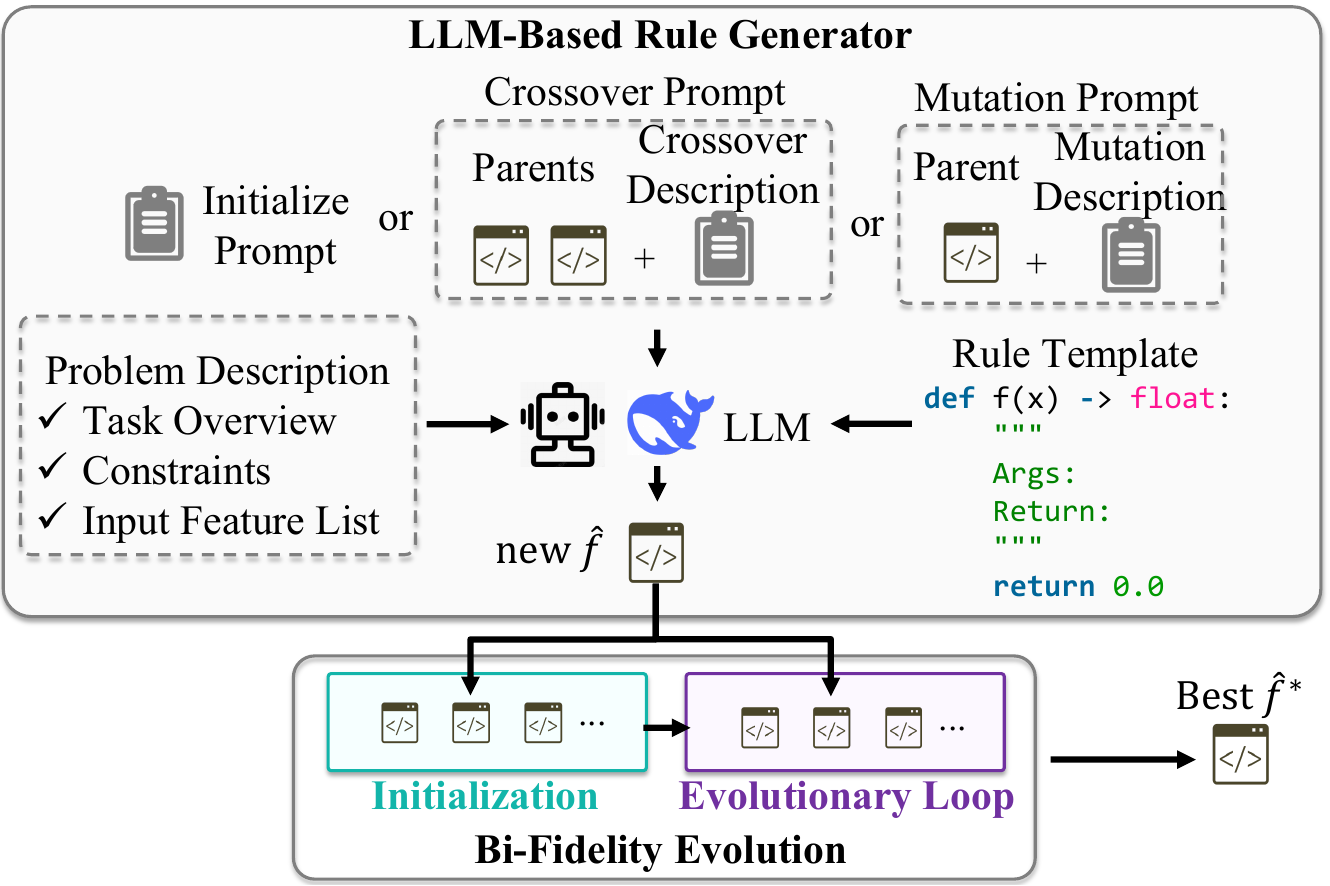}
    \caption{LLM-based rule generator. Given an operation instruction (initialization, crossover, or mutation), a problem description, and a rule template, the LLM generates a new rule. The LLM-generated rules undergo evolution and selection within the BiFE framework to yield an optimal rule.}
    \label{fig:llm_rule_generator}
    \vspace{-16pt}
\end{figure}

\subsection{LLM-Based Rule Generation}

We leverage an LLM to generate branching heuristics throughout the evolutionary process.
The LLM receives a structured prompt comprising: (1) a description of the input features and output semantics of a branching rule, (2) a code template, (3) an instruction indicating whether to initialize a diverse rule, perform crossover, or apply mutation, and (4) the selected parent rule(s), if applicable.
During initialization, no parents are available, so component (4) is omitted and the LLM generates the initial population from scratch.
In later generations, parents are provided and the instruction directs crossover or mutation accordingly.
The LLM directly outputs each new heuristic as code.
Heuristics that fail to execute (due to syntax errors, runtime errors, or timeout) are assigned a minimal fitness score and naturally eliminated during selection.
Details are provided in Appendix~\ref{appendix:implementation}.
The structure of the LLM-based rule generator is illustrated in Figure~\ref{fig:llm_rule_generator}.

\subsection{Elite-Prioritized Population Evolution}

The imperfect nature of $F_{LF}$ motivates retaining $F_{HF}$ as a fine-grained verifier to discover heuristics that surpass the imitation ceiling.
We define an elite population $\mathcal{E}$ as the subset of individuals that undergo high-fidelity evaluation.
In the initial generation, the LLM generates a population of heuristics from scratch, which are then ranked by $F_{LF}$, and the top $\alpha$ fraction form $\mathcal{E}$. Each $\hat{f} \in \mathcal{E}$ is then evaluated on real MILP instances to obtain $F_{HF}$.
In the main evolutionary loop, the elite admission threshold $\tau$ is set as the $F_{LF}$ of the individual with the worst $F_{HF}$ in $\mathcal{E}$,
$
\tau = F_{LF}\bigl(\arg\min_{\hat{f} \in \mathcal{E}} F_{HF}(\hat{f})\bigr),
$
which means that in subsequent generations, a candidate is admitted into $\mathcal{E}$ for high-fidelity evaluation provided that $F_{LF}(\hat{f}) \ge \tau$.

This hierarchical structure reshapes both parental and survival selection in the evolutionary loop, prioritizing the elite population in both forms of selection. For parent selection, individuals in $\mathcal{E}$ are ranked first by $F_{HF}$, followed by the general population ranked by $F_{LF}$. The combined pool is then sampled with probability proportional to rank, where the selection probability for the $r$-th individual is
$
p(r) \propto \frac{1}{r + N},
$
with $N$ denoting the overall population size. Selected parents are then passed to the LLM for crossover or mutation and producing new candidate.
Survival selection simply retains all members of $\mathcal{E}$, filling the remaining slots with the top-ranked general population individuals according to $F_{LF}$.

\begin{table*}[t]
\caption{The experimental results on standard benchmarks. Values are reported as geometric mean (geometric standard deviation) over 100 solving attempts per method. \textbf{Bold} font marks the best time among CPU-evaluated methods (our core focus), \underline{underline} marks cases where the best CPU method additionally outperforms all GPU-evaluated methods. Nodes are reported for completeness but not the primary evaluation metric, and $\star$ in the Nodes column marks cases that hit the 1000s time limit.}
\label{tab:standard_results}
\centering
\small
\setlength{\tabcolsep}{0.5mm}
\def\pm#1${\mathopen{(}#1\mathclose{)}$}
\textbf{Test Instances}\par\smallskip

\resizebox{\textwidth}{!}{%
\begin{tabular*}{485pt}{@{\extracolsep{\fill}}lcccccccc@{}}
\toprule
Method & \multicolumn{2}{c}{CA} & \multicolumn{2}{c}{SC} & \multicolumn{2}{c}{MIS} & \multicolumn{2}{c}{CFL} \\
\cmidrule(lr){2-3} \cmidrule(lr){4-5} \cmidrule(lr){6-7} \cmidrule(lr){8-9}
& Nodes & Time & Nodes & Time & Nodes & Time & Nodes & Time \\ \midrule
RPB       & $8.79 \pm 3.97$ & $2.31 \pm 1.60$ & $24.93 \pm 8.88$ & $6.13 \pm 1.75$ & $30.71 \pm 9.47$ & $8.74 \pm 1.72$ & $65.51 \pm 5.64$ & $33.70 \pm 2.07$ \\ \midrule
FiLM       & $57.68 \pm 2.93$ & $2.01 \pm 1.47$ & $87.23 \pm 4.00$ & $5.20 \pm 1.68$ & $108.52 \pm 10.57$ & $13.78 \pm 2.48$ & $228.93 \pm 3.17$ & $30.82 \pm 2.29$ \\
Symb4CO    & $63.10 \pm 3.06$ & $1.59 \pm 1.48$ & $114.74 \pm 4.37$ & $4.93 \pm 1.76$ & $70.00 \pm 7.07$ & $6.35 \pm 1.66$ & $264.45 \pm 3.12$ & $33.19 \pm 2.44$ \\
GS4CO      & $59.66 \pm 3.03$ & $1.93 \pm 1.43$ & $112.26 \pm 4.34$ & $5.21 \pm 1.70$ & $146.25 \pm 10.29$ & $8.66 \pm 2.03$ & $273.15 \pm 3.01$ & $40.22 \pm 2.36$ \\
GNN        & $59.41 \pm 3.09$ & $3.65 \pm 1.99$ & $92.33 \pm 3.66$ & $32.36 \pm 3.24$ & $51.74 \pm 6.16$ & $9.67 \pm 2.27$ & $217.83 \pm 3.03$ & $70.79 \pm 2.64$ \\
EoH        & $85.27 \pm 3.15$ & $1.62 \pm 1.45$ & $113.36 \pm 4.17$ & $4.73 \pm 1.71$ & $674.73 \pm 30.81$ & $22.61 \pm 5.83$ & $265.82 \pm 3.04$ & $28.37 \pm 2.25$ \\
ReEvo      & $817.17 \pm 6.21$ & $4.58 \pm 3.22$ & $109.00 \pm 4.18$ & $4.71 \pm 1.70$ & $89.70 \pm 7.14$ & $6.44 \pm 1.59$ & $272.33 \pm 3.03$ & $29.60 \pm 2.25$ \\
LLM4Branch & $80.14 \pm 2.88$ & $1.64 \pm 1.26$ & $148.95 \pm 4.15$ & $4.77 \pm 1.52$ & $93.93 \pm 6.71$ & $6.57 \pm 1.43$ & $281.54 \pm 3.01$ & $31.11 \pm 2.18$ \\
BiFE       & $70.50 \pm 3.11$ & $\textbf{\underline{1.51}} \pm 1.41$ & $109.17 \pm 4.10$ & $\textbf{\underline{4.58}} \pm 1.67$ & $81.75 \pm 7.75$ & $\textbf{\underline{6.26}} \pm 1.62$ & $266.52 \pm 2.99$ & $\textbf{\underline{28.29}} \pm 2.22$ \\ \midrule
GNN-GPU    & $59.41 \pm 3.09$ & $1.93 \pm 1.48$ & $92.33 \pm 3.66$ & $5.07 \pm 1.71$ & $51.74 \pm 6.16$ & $6.96 \pm 1.53$ & $217.83 \pm 3.03$ & $29.23 \pm 2.18$ \\
tMDP-GPU   & $94.90 \pm 2.81$ & $2.09 \pm 1.50$ & $406.79 \pm 5.82$ & $10.13 \pm 2.94$ & $119.46 \pm 8.87$ & $7.96 \pm 2.17$ & $381.80 \pm 3.74$ & $43.60 \pm 2.76$ \\ \bottomrule
\end{tabular*}}

\medskip

\textbf{Transfer Instances}\smallskip

\resizebox{\textwidth}{!}{%
\begin{tabular*}{485pt}{@{\extracolsep{\fill}}lcccccccc@{}}
\toprule
Method & \multicolumn{2}{c}{CA} & \multicolumn{2}{c}{SC} & \multicolumn{2}{c}{MIS} & \multicolumn{2}{c}{CFL} \\
\cmidrule(lr){2-3} \cmidrule(lr){4-5} \cmidrule(lr){6-7} \cmidrule(lr){8-9}
& Nodes & Time & Nodes & Time & Nodes & Time & Nodes & Time \\ \midrule
RPB       & $766.90 \pm 4.16$ & $16.80 \pm 1.59$ & $3624.9 \pm 3.83$ & $\textbf{\underline{68.25}} \pm 2.16$ & $2701.3 \pm 7.55$ & $115.67 \pm 2.49$ & $144.29 \pm 9.26$ & $99.42 \pm 2.94$ \\ \midrule
FiLM       & $848.24 \pm 2.74$ & $16.23 \pm 2.00$ & $3090.0 \pm 3.36$ & $70.56 \pm 2.61$ & $1552.4 \pm 12.35$ & $112.89 \pm 5.16$ & $488.62 \pm 5.11$ & $100.99 \pm 3.38$ \\
Symb4CO    & $933.03 \pm 2.77$ & $15.37 \pm 2.19$ & $4883.4 \pm 3.53$ & $85.12 \pm 2.88$ & $3880.1 \pm 9.99$ & $176.70 \pm 4.42$ & $504.12 \pm 5.16$ & $105.92 \pm 3.61$ \\
GS4CO      & $1042.5 \pm 3.04$ & $14.40 \pm 2.03$ & $4429.2 \pm 3.50$ & $76.38 \pm 2.72$ & $16044^\star \pm 14.84$ & $452.49 \pm 6.99$ & $464.10 \pm 4.65$ & $110.36 \pm 3.07$ \\
GNN        & $884.32 \pm 2.78$ & $54.95 \pm 2.57$ & $2217.6^{\star} \pm 2.42$ & $771.15 \pm 1.72$ & $1411.6^{\star} \pm 7.83$ & $200.77 \pm 4.77$ & $420.31^\star \pm 4.82$ & $248.22 \pm 4.01$ \\
EoH        & $1261.1 \pm 2.87$ & $13.57 \pm 2.05$ & $4067.4 \pm 3.37$ & $77.69 \pm 2.82$ & $22914^\star \pm 5.26$ & $587.86 \pm 3.18$ & $479.60 \pm 4.88$ & $84.30 \pm 3.11$ \\
ReEvo      & $87523^\star \pm 3.35$ & $446.89 \pm 3.14$ & $3976.0 \pm 3.37$ & $76.79 \pm 2.77$ & $2354.0 \pm 6.77$ & $89.99 \pm 3.16$ & $495.37 \pm 4.89$ & $89.55 \pm 3.16$ \\
LLM4Branch & $1243.1 \pm 2.68$ & $13.95 \pm 1.74$ & $5618.5 \pm 3.53$ & $69.78 \pm 2.72$ & $2963.5 \pm 7.30$ & $90.97 \pm 3.05$ & $507.17 \pm 4.93$ & $96.19 \pm 3.09$ \\
BiFE       & $1092.7 \pm 2.68$ & $\textbf{\underline{11.95}} \pm 1.92$ & $3820.8 \pm 3.33$ & $71.80 \pm 2.73$ & $2547.4 \pm 5.98$ & $\textbf{89.07} \pm 2.98$ & $474.91 \pm 4.69$ & $\textbf{\underline{83.89}} \pm 2.99$ \\ \midrule
GNN-GPU    & $884.32 \pm 2.78$ & $16.02 \pm 2.03$ & $2815.9 \pm 3.27$ & $69.29 \pm 2.63$ & $1430.2 \pm 8.03$ & $82.95 \pm 3.23$ & $435.50 \pm 5.16$ & $96.88 \pm 3.22$ \\
tMDP-GPU   & $1768.5 \pm 3.08$ & $22.70 \pm 2.26$ & $43880^\star \pm 3.87$ & $735.99 \pm 3.70$ & $4085.3^\star \pm 9.05$ & $113.87 \pm 4.21$ & $788.73^\star \pm 5.44$ & $139.43 \pm 3.77$ \\ \bottomrule
\end{tabular*}}

\vspace{-10pt}
\end{table*}

\section{Experiments}
\subsection{Settings}
\noindent\textbf{Benchmarks.}
Consistent with prior studies~\citep{gasse2019exact,gupta2020hybrid,scavuzzo2022tmdp,kuang2024rethinking,kuang2024towards}, we evaluate BiFE on four standard benchmark problems: Combinatorial Auctions (CA), Set Covering (SC), Maximum Independent Set (MIS), and Capacitated Facility Location (CFL). For each problem, we assess the generated branching rules on two categories of instances: those of the same scale as the instances used during the algorithm design phase (referred to as \enquote{test instances}) and those of a larger scale (referred to as \enquote{transfer instances}).
To validate the performance of BiFE on real-world problems, we further evaluate on three benchmark problems drawn from the NeurIPS 2021 ML4CO competition~\citep{gasse2022ml4co} and the Distributional MIPLIB library~\citep{huang2024dmiplib}: Balanced Item Placement (BIP), Neural Network Verification (NNV), and Optimal Transmission Switching (OTS).
The specific sizes of the test and transfer instances for the standard benchmarks, along with further details regarding the three real-world benchmarks, are documented in Appendix~\ref{appendix:benchmarks}.
\noindent\textbf{Evaluation Metrics.}
\label{section:evaluation_metrics}
We report the number of B\&B nodes and solving time for standard benchmarks and real-world benchmarks NNV and OTS. Since our ultimate goal is to minimize solving time, a metric determined by both node count and per-node inference cost, fewer nodes do not guarantee faster solving if inference is expensive. Thus, solving time serves as the primary metric, with node count provided for reference.
For BIP, instances are extremely computationally challenging and measuring solving time or node counts is not practical.
Following common practice for such cases~\citep{gasse2022ml4co}, we evaluate BIP using the primal-dual integral (PDI) and the optimality gap.
The PDI quantifies the area enclosed by the solver's primal and dual bound curves over time, thus rewarding both progress in finding good feasible solutions and in improving the dual bound.
Given a time limit $T$, it is defined as
$
    \text{PDI} = \int_{t=0}^{T} \bigl( \mathbf{c}^\top \mathbf{x}_t^\star - \mathbf{y}_t^\star \bigr) \, \mathrm{d}t,
$
where $\mathbf{x}_t^\star$ is the best feasible solution up to time $t$, and $\mathbf{y}_t^\star$ is the best dual bound at time $t$.
The optimality gap is defined as the relative gap between the best primal and dual bounds at the end of the time limit $T$:
$
    \text{gap} = |\mathbf{c}^\top \mathbf{x}_T^\star - \mathbf{y}_T^\star| / (|\mathbf{c}^\top \mathbf{x}_T^\star| + \epsilon),
$
where $\epsilon$ is a small constant for numerical stability.

\noindent\textbf{Baselines.}
We evaluate against four categories of baselines:
(1) Handcrafted heuristic: SCIP's default branching rule, reliable pseudo-cost branching (RPB)~\citep{scip}, represents a highly engineered heuristic adopted in solvers.
(2) CPU-efficient policies: FiLM~\citep{gupta2020hybrid} uses a lightweight neural architecture, while Symb4CO~\citep{kuang2024rethinking} and GS4CO~\citep{kuang2024towards} employ neuro-symbolic search to obtain symbolic policies.
(3) General neural policies: GNN~\citep{gasse2019exact} (imitation learning) and tMDP~\citep{scavuzzo2022tmdp} (reinforcement learning). The \enquote{-GPU} suffix denotes GPU evaluation and default denotes CPU.
(4) LLM-based algorithm design: EoH~\citep{liu2024EoH} and ReEvo~\citep{ye2024reevo} are general-purpose LLM-based frameworks that directly use low-fidelity $L_{LF}$ to evolve heuristics. LLM4Branch~\citep{hou2026llm4branch} generates program skeletons and tunes their numerical parameters with direct solver feedback. Together, they contrast low-fidelity-only evolution and repeated high-fidelity solver tuning with BiFE's bi-fidelity evaluation allocation.
We note that other LLM-based methods targeting different MILP components are not directly comparable within the branching policy evaluation framework.
A detailed discussion of these methods is provided in Appendix~\ref{appendix:other}.

\noindent\textbf{Implementation.}
\label{section:implementation}
We implement BiFE and all baselines within SCIP 7.0.3~\citep{scip}. Following prior work~\citep{gasse2019exact,gupta2020hybrid,kuang2024rethinking,kuang2024towards}, we modify only the branching rule, allow cutting plane generation at the root node only, and deactivate solver restarts, with all other SCIP parameters kept at their default values. For evaluation, all methods are tested on 20 instances across 5 random seeds, totaling 100 solving attempts, with a time limit of 1,000 seconds per attempt. All evaluations are conducted on identical hardware.
For LLM-based methods, we use DeepSeek-V4-Flash~\citep{deepseek} as the backbone model with consistent evolutionary parameters (such as number of LLM calls and population size). 
Prompt design (crossover and mutation, problem descriptions, etc.) and other details are provided in Appendix~\ref{appendix:implementation}.

\noindent\textbf{BiFE Setup.}
To construct $F_{LF}$, we collect 1,000 strong-branching samples per benchmark, and $F_{HF}$ is evaluated on 20 disjoint instances. For $F_{HF}$, the performance metric varies by benchmark: for CA, SC, MIS, and CFL, it is solving time with a 60-second per-instance time limit. For NNV, OTS, and BIP, the time limit is 300 seconds, with metrics being solving time, optimality gap, and PDI, respectively. 
For a fair comparison, LLM4Branch uses comparable time limits across its three evaluation stages. 
Full per-benchmark settings are provided in Appendix~\ref{appendix:implementation}.

\begin{table*}[t]
\caption{Experimental results on real-world benchmarks. Similarly, values are reported as geometric mean (geometric standard deviation) over 100 solving attempts per method. \textbf{Bold} marks the best gap/PDI/time among CPU methods, \underline{underline} indicates the best CPU method outperforms all GPU methods. $\star$ indicates hitting the 1000s time limit.}
\label{tab:realworld_results}
\centering
\scriptsize
\setlength{\tabcolsep}{0.5mm}
\def\pm#1${\mathopen{(}#1\mathclose{)}$}
\begin{tabular*}{\textwidth}{@{\extracolsep{\fill}}lcccccc@{}}
\toprule
Method & \multicolumn{2}{c}{BIP} & \multicolumn{2}{c}{OTS} & \multicolumn{2}{c}{NNV} \\
\cmidrule(lr){2-3} \cmidrule(lr){4-5} \cmidrule(lr){6-7}
& PDI & Gap & Nodes & Time & Nodes & Time \\ \midrule
RPB    & $81057.56 \pm 1.24$ & $93.47 \pm 13.99$ & $939.91 \pm 9.01$ & $320.62 \pm 2.60$ & $372.96 \pm 31.85$ & $22.96 \pm 7.85$ \\ \midrule
Symb4CO & $89220.25 \pm 1.07$ & $132.48 \pm 11.95$ & $1474.2^\star \pm 3.17$ & $544.89 \pm 1.94$ & $365.89 \pm 60.40$ & $25.43 \pm 15.50$ \\
GS4CO   & $87270.12 \pm 1.12$ & $118.91 \pm 12.09$ & $2107.4^\star \pm 2.11$ & $961.48 \pm 1.16$ & $502.99 \pm 80.20$ & $25.55 \pm 18.31$ \\
GNN     & $87835.06 \pm 1.08$ & $88.39 \pm 6.29$ & $269.80^\star \pm 2.96$ & $479.07 \pm 2.42$ & $254.01^\star \pm 18.54$ & $58.55 \pm 11.49$ \\
LLM4Branch & $\textbf{\underline{76010.83}} \pm 1.30$ & $72.25 \pm 14.12$ & $4709.2^\star \pm 2.29$ & $785.75 \pm 1.51$ & $362.27 \pm 23.67$ & $18.32 \pm 4.49$ \\
BiFE    & $76152.79 \pm 1.21$ & $\textbf{65.85} \pm 16.13$ & $4327.6 \pm 3.76$ & $\textbf{306.28} \pm 2.40$ & $242.54 \pm 18.69$ & $\textbf{\underline{12.58}} \pm 5.22$ \\
\midrule
GNN-GPU & $82599.38 \pm 1.11$ & $52.67 \pm 3.89$ & $561.03 \pm 6.61$ & $211.45 \pm 2.67$ & $270.85 \pm 20.06$ & $20.55 \pm 6.60$ \\
\bottomrule
\end{tabular*}

\vspace{-6pt}
\end{table*}

\subsection{Results}
\noindent\textbf{Standard Benchmarks.}
The experimental results in Table~\ref{tab:standard_results} demonstrate the effectiveness of BiFE across all standard benchmarks. On test instances, BiFE achieves the best solving time among all CPU-evaluated methods across all four benchmarks, and these results also surpass GPU-evaluated baselines. 
LLM4Branch provides a closely competitive LLM-generated-code baseline on these distributions.
On transfer instances involving larger problem scales, BiFE maintains its leading position among CPU-evaluated methods on CA, MIS, and CFL, with CA and CFL again outperforming GPU-evaluated baselines. LLM4Branch edges ahead only on SC transfer, where both still trail RPB.
Notably, BiFE achieves shorter solving time while exploring more nodes than certain baselines. Since solving time is the primary metric, node count alone does not fully reflect branching policy performance. The reasons behind this phenomenon are further examined in Discussion. More statistical tests are reported in Appendix~\ref{appendix:statistical_tests}.

\noindent\textbf{Real-World Benchmarks.}
The results on real-world benchmarks are presented in Table~\ref{tab:realworld_results}, where we evaluate a subset of baselines that are competitive on the standard benchmarks.
On BIP, LLM4Branch obtains a 0.19\% lower PDI, while BiFE yields the lowest gap among CPU-evaluated methods, indicating comparable solving efficiency between the two approaches.
On NNV and OTS, BiFE achieves the best solving time among CPU-evaluated methods on both benchmarks, outperforming LLM4Branch and even surpassing GNN-GPU on NNV. Notably, BiFE also attains the fewest nodes on NNV among all methods.

\subsection{Discussion}
\label{section:discussion}
\begin{figure*}[t]
    \centering
    \includegraphics[height=0.225\textwidth]{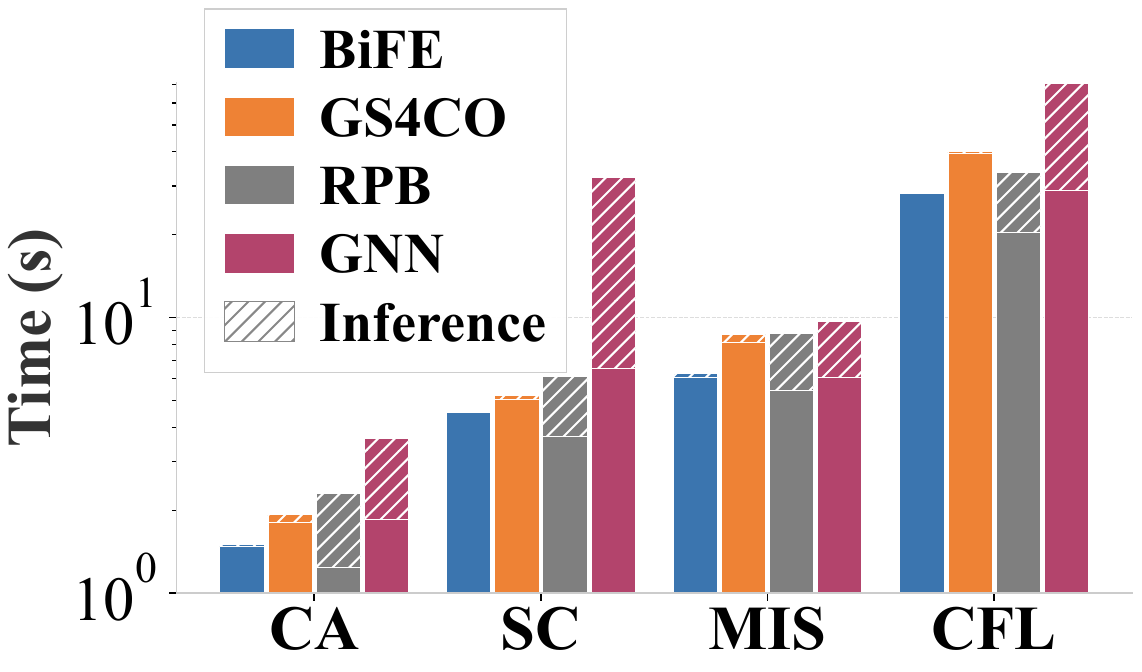}\hfill
    \includegraphics[height=0.25\textwidth]{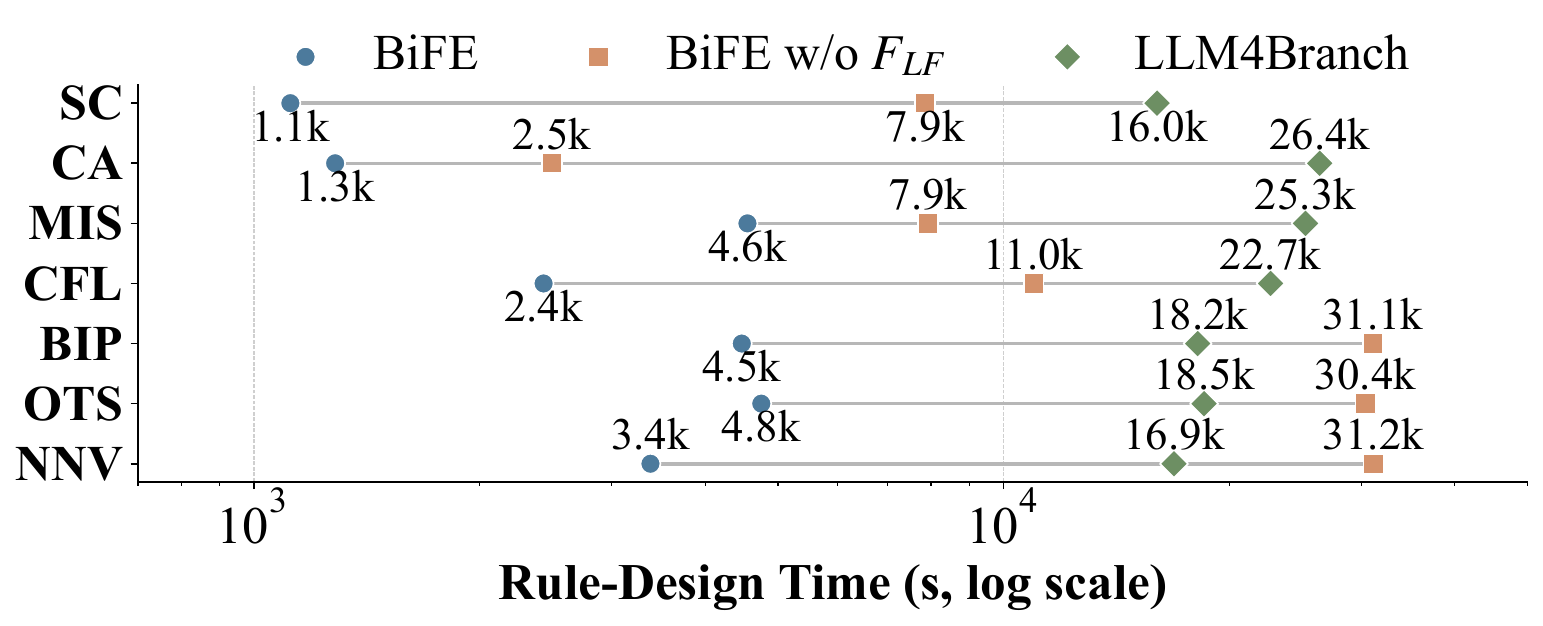}
    \caption{(a) Left:~Inference time in total time of BiFE, RPB, GS4CO, and GNN on test instances of standard benchmarks. (b) Right:~Rule-design time across benchmarks. BiFE includes evolutionary search and the one-off cost of collecting strong-branching demonstrations. LLM4Branch's time includes solver-based Bayesian parameter optimization and validation.}
    \label{fig:combined_analysis}
\vspace{-6pt}
\end{figure*}

\noindent\textbf{More Nodes, Yet Less Time.}
Tables~\ref{tab:standard_results} and~\ref{tab:realworld_results} show that RPB and neural policies (GNN, FiLM) explore fewer nodes than neuro-symbolic (Symb4CO, GS4CO) and LLM-based methods (BiFE, EoH, ReEvo, LLM4Branch), yet often incur longer solving times. As Figure~\ref{fig:combined_analysis}(a) reveals, this stems from their high inference cost: RPB relies on expensive strong branching for initialization, while neural methods are inefficient on CPU, unlike on GPU. Neuro-symbolic and LLM-based methods, by contrast, use lightweight symbolic or LLM-generated rules with minimal CPU overhead, exploring more nodes but solving faster. Among them, Symb4CO and GS4CO sometimes underperform BiFE despite similar or fewer nodes, as their official implementations combine RPB with symbolic rules, raising inference cost. BiFE avoids this by relying solely on lightweight rules, achieving the best time efficiency.

\begin{figure*}[t]
    \centering
    \includegraphics[width=\linewidth]{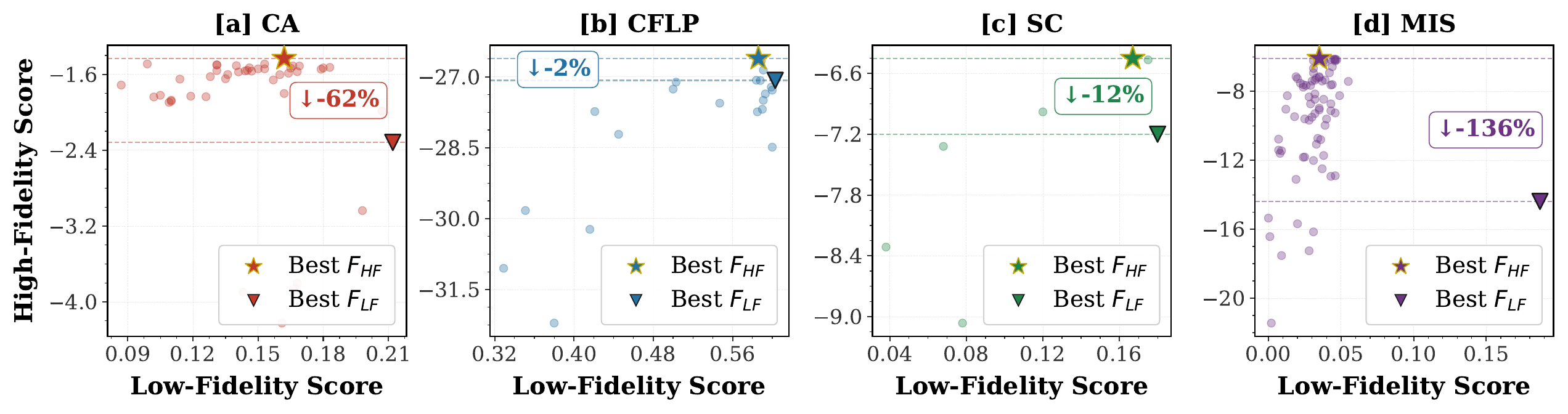}
    \caption{Scatter plot of $F_{LF}$ vs. $F_{HF}$ highlighting the performance drop based solely on $F_{LF}$.}
    \label{fig:lf_hf_correlation}
\vspace{-6pt}
\end{figure*}

\noindent\textbf{Limitations of Relying Solely on $F_{LF}$.}
We observe that EoH and ReEvo perform comparably to BiFE on most benchmarks, yet fail drastically on specific ones---EoH on MIS and ReEvo on CA (see Table~\ref{tab:standard_results}). To investigate this, we visualize the relationship between $F_{LF}$ and $F_{HF}$ using all individuals from BiFE's populations on standard benchmarks.
Figure~\ref{fig:lf_hf_correlation} presents the scatter plots highlighting the performance drop from the best $F_{LF}$ to the true best $F_{HF}$ on each benchmark. The degradation is most severe on MIS and CA, precisely mirroring the failure cases of EoH and ReEvo.
This directly validates our core argument: due to distribution shift, $F_{LF}$ based on offline datasets cannot fully substitute for $F_{HF}$.
The imperfect $F_{LF}$ thus necessitates retaining $F_{HF}$ as a fine-grained verifier within our bi-fidelity framework.

\noindent\textbf{Search Efficiency Relative to LLM4Branch.}
The performance results in Tables~\ref{tab:standard_results} and~\ref{tab:realworld_results} indicate largely comparable solution quality: BiFE matches or slightly exceeds LLM4Branch on the standard benchmarks and BIP, and is stronger on OTS and NNV.
Despite this comparable solution quality, Figure~\ref{fig:combined_analysis}(b) reports a pronounced difference in rule-design time. To ensure a fair comparison, BiFE's total budget includes evolutionary search and the one-off cost of collecting strong-branching samples to construct $F_{LF}$, while LLM4Branch is evaluated under time limits aligned with BiFE.
Under these settings, BiFE requires 3.9$\times$ to 20.6$\times$ less time than LLM4Branch across all benchmarks.
This marked reduction reflects their different evaluation workflows: LLM4Branch runs fitness evaluation both before and after Bayesian search, each driven by costly solver feedback.
By screening the population with $F_{LF}$ and reserving $F_{HF}$ for elite candidates, BiFE avoids repeated solver-feedback workload. The resulting order-of-magnitude reduction shows that BiFE can attain comparable downstream performance while spending far less time on rule discovery, demonstrating a substantial improvement in search efficiency.


\subsection{Ablation Study}

\begin{table}[t]
\small
\centering
\begin{tabular}{lcccc}
\toprule
Method & SC & CA & CFL & MIS \\
\midrule
BiFE w/o $F_{LF}$ & 4.66 & 1.52 & 29.38 & 6.54 \\
BiFE              & 4.58 & 1.51 & 28.29 & 6.26 \\
\bottomrule
\end{tabular}
\caption{Times of rules designed by BiFE and BiFE w/o $F_{LF}$ on the test instances of standard benchmarks.}
\label{tab:ablation}
\vspace{-12pt}
\end{table}

We construct an ablation study to investigate whether BiFE's bi-fidelity framework indeed offers superior search efficiency compared to relying solely on $F_{HF}$. To isolate this effect, we compare BiFE against a variant that removes the low-fidelity filter, denoted BiFE w/o $F_{LF}$, which directly evaluates all candidates via $F_{HF}$. Both configurations are run with identical generations and population sizes across all benchmarks, and BiFE's reported time retains the one-off $F_{LF}$ data-collection cost. The rule-design time of both methods is shown in Figure~\ref{fig:combined_analysis}(b).
Furthermore, we compare the solving times of rules designed by both methods on the test instances of standard benchmarks, presented in Table~\ref{tab:ablation}.

As shown, BiFE achieves substantially shorter rule-design time than BiFE w/o $F_{LF}$ across all benchmarks, with speedups ranging from 1.7$\times$ on MIS to over 9.2$\times$ on NNV, even with the offline data collection overhead included.
Meanwhile, the rules designed by both methods show no significant difference in performance.
This confirms that the low-fidelity pre-screener effectively prunes the search space, enabling the bi-fidelity framework to discover branching rules with far higher search efficiency than pure high-fidelity evaluation, without compromising the quality of the resulting rules.

\section{Conclusion}
In this paper, we introduce BiFE, a bi-fidelity evolutionary framework that discovers LLM-generated branching policies for MILP. By using low-fidelity imitation scores as a fast pre-screener and reserving expensive on-instance evaluation for elite candidates, BiFE effectively balances search efficiency with performance reliability. Experiments on standard and real-world benchmarks show that the discovered rules outperform SCIP's default heuristic, existing CPU-efficient baselines, and even certain GPU-based neural policies, demonstrating the promise of LLM-generated code as a lightweight, interpretable, and expressive representation for branching.
However, the policies discovered by BiFE remain benchmark-specific. In future work, we will focus on LLM-designed branching rules that generalize across instances, rather than 
policies tailored to a single benchmark.

\section*{Limitations}
While BiFE effectively discovers CPU-efficient and complex branching policies within the LLM-generated code space, the discovered policies remain benchmark-specific. Currently, the evolutionary search is conducted and tailored independently for each problem distribution (e.g., CA, SC, MIS). Consequently, a policy evolved for one benchmark may not generalize well to structurally different MILP instances, necessitating the search process to be rerun for new problem domains. In future work, we aim to focus on LLM-designed branching rules that generalize across instances, exploring universal algorithmic logic that can adapt to or be shared among diverse MILP benchmarks, rather than relying on benchmark-specific policies.



\bibliography{custom}

\newpage
\appendix

\section{Algorithm Pseudocode of BiFE}
\label{appendix:pseudocode}

A more detailed algorithmic description of the BiFE framework introduced in the main paper is provided in Algorithm~\ref{alg:bife}, covering initialization, elite-prioritized evolution, and survival selection.

\begin{algorithm}[h]
\caption{BiFE: Bi-Fidelity Evolutionary Framework}
\label{alg:bife}
\SetKwInOut{Input}{Input}
\SetKwInOut{Output}{Output}
\Input{LLM generator $\mathcal{G}$, offline $\mathcal{D}$, MILP instance set $\mathcal{M}$, population size $N$, elite fraction $\alpha$, generations $K$}
\Output{Best heuristic $\hat{f}^\star$}

\BlankLine
$\mathcal{P}_0 \gets \mathcal{G}.\text{initialize}(N)$\;
\For{$\hat{f} \in \mathcal{P}_0$}{
    Evaluate $F_{LF}(\hat{f})$ on $\mathcal{D}$\;
}
Sort $\mathcal{P}_0$ by $F_{LF}$ descending\;
$\mathcal{E} \gets \text{top } \lceil \alpha N \rceil \text{ of } \mathcal{P}_0$\;
\For{$\hat{f} \in \mathcal{E}$}{
    Evaluate $F_{HF}(\hat{f})$ on $\mathcal{M}$\;
}
$\tau \gets F_{LF}(\arg\min_{\hat{f} \in \mathcal{E}} F_{HF}(\hat{f}))$\;

\BlankLine
\For{$k \gets 1$ \KwTo $K$}{
    $\mathcal{P}_{\text{new}} \gets \emptyset$\;
    \For{each crossover and mutation operator}{
        Select parent pair $(\hat{f}_a, \hat{f}_b)$ via elite-prioritized ranking\;
        $\hat{f}_{\text{child}} \gets \mathcal{G}.\text{crossover}(\hat{f}_a, \hat{f}_b)$ or $\mathcal{G}.\text{mutation}(\hat{f}_a)$\;
        Evaluate $F_{LF}(\hat{f}_{\text{child}})$ on $\mathcal{D}$\;
        \If{$F_{LF}(\hat{f}_{\text{child}}) \ge \tau$}{
            Evaluate $F_{HF}(\hat{f}_{\text{child}})$ on $\mathcal{M}$\;
            $\mathcal{E} \gets \mathcal{E} \cup \{\hat{f}_{\text{child}}\}$\;
        }
        $\mathcal{P}_{\text{new}} \gets \mathcal{P}_{\text{new}} \cup \{\hat{f}_{\text{child}}\}$\;
    }
    Update $\tau$ w.r.t.\ current $\mathcal{E}$\;
    $\mathcal{P}_k \gets \mathcal{E} \cup \text{top } (N - |\mathcal{E}|) \text{ of } (\mathcal{P}_{\text{new}} \setminus \mathcal{E}) \text{ by } F_{LF}$\;
}
\Return $\arg\max_{\hat{f} \in \mathcal{E}} F_{HF}(\hat{f})$\;
\end{algorithm}

\section{More Details on Benchmarks}
\label{appendix:benchmarks}

For each standard benchmark, the parameter settings in the test and transfer instances are presented in Table~\ref{tab:instance}. Notably, the algorithm design and testing phases employ instances of the same size. Additionally, to further evaluate the generalization capability of the generated branching policy, transfer instances of a larger scale have been specifically designed.

The Balanced Item Placement (BIP) problem involves spreading items, e.g., files or processes, across containers, e.g., disks or machines, utilizing them evenly. Items can have multiple copies, but at most one copy can be placed in a single bin. The number of items that can be moved is constrained, modeling the real-life situation of a live system for which some placement already exists. In the ML4CO competition~\citep{gasse2022ml4co}, this dataset contains 10,000 training instances (pre-split into 9,900 train and 100 validation instances) and 100 test instances.

The Neural Network Verification (NNV) problem addresses the verification of adversarial robustness for a pre-trained neural network on a given input example. The goal is to determine the minimum perturbation to the input required to change the network's output from the correct class to an incorrect one; if this minimum perturbation exceeds a specified robustness threshold, the network is considered robust on that input. The NNV instances we use belong to the \enquote{easy} hardness level in the Distributional MIPLIB~\citep{huang2024dmiplib}.

The Optimal Transmission Switching (OTS) problem arises in energy planning under high wildfire ignition risk. The transmission grid is modeled as a graph of power buses (vertices) connected by transmission lines (edges). During high-risk conditions, lines may ignite wildfires; the problem seeks the optimal strategy to de-energize or underground transmission lines to mitigate wildfire risk while minimizing power outages under a resource budget. The OTS instances we use belong to the \enquote{easy} hardness level in the Distributional MIPLIB~\citep{huang2024dmiplib}.

\begin{table}[t]
\caption{Size of the test and transfer instances in standard benchmarks}
\label{tab:instance}
\centering
\begin{tabular}{@{}cccc@{}}
\toprule
Benchmarks                                                              & Parameters                                                     & Test                                              & Transfer                                           \\ \midrule
\begin{tabular}[c]{@{}c@{}}Set\\ Covering\end{tabular}                  & \begin{tabular}[c]{@{}c@{}}Items\\ Sets\end{tabular}           & \begin{tabular}[c]{@{}c@{}}500\\ 1000\end{tabular} & \begin{tabular}[c]{@{}c@{}}1000\\ 1000\end{tabular} \\ \midrule
\begin{tabular}[c]{@{}c@{}}Combinatorial\\ Auctions\end{tabular}        & \begin{tabular}[c]{@{}c@{}}Items\\ Bids\end{tabular}           & \begin{tabular}[c]{@{}c@{}}100\\ 500\end{tabular} & \begin{tabular}[c]{@{}c@{}}200\\ 1000\end{tabular}  \\ \midrule
\begin{tabular}[c]{@{}c@{}}Maximum\\ Independent Set\end{tabular}       & \begin{tabular}[c]{@{}c@{}}Nodes\\ Affinity\end{tabular}       & \begin{tabular}[c]{@{}c@{}}500\\ 4\end{tabular}   & \begin{tabular}[c]{@{}c@{}}1000\\ 4\end{tabular}    \\ \midrule
\begin{tabular}[c]{@{}c@{}}Capacitated\\ Facility Location\end{tabular} & \begin{tabular}[c]{@{}c@{}}Customers\\ Facilities\end{tabular} & \begin{tabular}[c]{@{}c@{}}100\\ 100\end{tabular}   & \begin{tabular}[c]{@{}c@{}}200\\ 100\end{tabular}    \\ \bottomrule
\end{tabular}

\end{table}

\begin{table*}[ht]
\caption{Comparison between BiFE and STRCMP on standard benchmarks.}
\label{tab:bife_strcmp}
\centering
\small
\setlength{\tabcolsep}{0.5mm}
\let\standardpm\pm
\renewcommand{\pm}{\mathord{\standardpm}}
{\centering\bfseries Test Instances\\[0.5em]}

\resizebox{\textwidth}{!}{%
\begin{tabular*}{452pt}{@{\extracolsep{\fill}}lcccccccc@{}}
\toprule
Method & \multicolumn{2}{c}{CA} & \multicolumn{2}{c}{SC} & \multicolumn{2}{c}{MIS} & \multicolumn{2}{c}{CFL} \\
\cmidrule(lr){2-3} \cmidrule(lr){4-5} \cmidrule(lr){6-7} \cmidrule(lr){8-9}
& Nodes & Time & Nodes & Time & Nodes & Time & Nodes & Time \\ \midrule
BiFE   & $70.50 \pm 3.11$ & $1.51 \pm 1.41$ & $109.17 \pm 4.10$ & $4.58 \pm 1.67$ & $81.75 \pm 7.75$ & $6.26 \pm 1.62$ & $266.52 \pm 2.99$ & $28.29 \pm 2.22$ \\
STRCMP & $8.88 \pm 3.77$ & $2.69 \pm 1.52$ & $23.06 \pm 8.46$ & $7.20 \pm 1.73$ & $25.93 \pm 9.76$ & $4.81 \pm 2.03$ & $70.86 \pm 5.77$ & $32.61 \pm 2.03$ \\ \bottomrule
\end{tabular*}}

\vspace{0.75em}

{\centering\bfseries Transfer Instances\\[0.5em]}

\resizebox{\textwidth}{!}{%
\begin{tabular*}{452pt}{@{\extracolsep{\fill}}lcccccccc@{}}
\toprule
Method & \multicolumn{2}{c}{CA} & \multicolumn{2}{c}{SC} & \multicolumn{2}{c}{MIS} & \multicolumn{2}{c}{CFL} \\
\cmidrule(lr){2-3} \cmidrule(lr){4-5} \cmidrule(lr){6-7} \cmidrule(lr){8-9}
& Nodes & Time & Nodes & Time & Nodes & Time & Nodes & Time \\ \midrule
BiFE   & $1092.66 \pm 2.68$ & $11.95 \pm 1.92$ & $3820.80 \pm 3.33$ & $71.80 \pm 2.73$ & $2547.39 \pm 5.98$ & $89.07 \pm 2.98$ & $474.91 \pm 4.69$ & $83.89 \pm 2.99$ \\
STRCMP & $752.74 \pm 4.26$ & $18.34 \pm 1.47$ & $3767.86 \pm 3.76$ & $64.53 \pm 2.13$ & $2329.81 \pm 7.68$ & $83.74 \pm 2.83$ & $149.88 \pm 9.88$ & $86.78 \pm 3.15$ \\ \bottomrule
\end{tabular*}}

\end{table*}

\section{Discussion of Related LLM-based Methods for MILP}
\label{appendix:other}

Several recent works apply LLM-based algorithm design to other components of the MILP solve process. Here we clarify their relationship to our work and explain why they are not included in the main comparison.

\subsection{Primal heuristics} 
Methods such as~\citet{zhou2024llmsolver,zhang2026dhevo,ye2025llm4lns} use LLMs to generate primal heuristics, which aim to find feasible solutions quickly during the solving process. Unlike branching policies and cutting plane selection, primal heuristics do not operate within the exact optimization framework: they only produce feasible solutions without optimality guarantees. We therefore exclude primal heuristic methods from our evaluation.

\subsection{Cutting plane selection} 
Several works apply LLM-based design to cutting plane selection and generation, including STRCMP~\citep{li2026strcmp} and EvoCut~\citep{yazdani2025evocut}. We note that cutting plane selection and branching policies are entirely orthogonal stages of the branch-and-cut framework and can in principle be jointly applied. Since both operate within the exact optimization paradigm, unlike primal heuristics, a comparison is at least not categorically ruled out. We select STRCMP as a reference point as it is evaluated on benchmarks comparable to ours (e.g., CA, SC, MIS, CFL), whereas EvoCut targets a different set of problems (e.g., TSP, JSSP). We provide a brief comparison with STRCMP on standard benchmarks in Table~\ref{tab:bife_strcmp}, noting that the comparison cannot be strictly controlled: STRCMP uses RPB for branching and its designed method for cutting plane selection, whereas we use our designed branching policy and default cutting planes.
Across both test and transfer instances, BiFE achieves lower solving times on the majority of benchmarks (five out of eight). While the comparison is not fully controlled, BiFE compares favorably to STRCMP overall.

\subsection{Variable selection rules scheduling}
Zheng et al.~\citeyearpar{zheng2026llm4scheduling} propose using LLMs to dynamically schedule handcrafted branching rules online, rather than discovering new rules. As the method is not open-sourced and targets a different problem formulation, we do not include it in our comparison.

\section{Implementation Details}
\label{appendix:implementation}

\subsection{LLM Generation Details}
Heuristics that raise syntax errors, runtime exceptions, or exceed the 30-second execution timeout are assigned a minimal fitness score and naturally eliminated during selection without explicit repair. To encourage computational efficiency, the prompt instructs the LLM to design lightweight rules; external dependencies are restricted to \texttt{numpy} and \texttt{math}, which is pre-imported in the rule template. Heuristics with excessive overhead are further penalized by poor solving time under $F_{HF}$, leading to automatic elimination.

\subsection{Hardware and environment}
All experiments are conducted on a server equipped with Intel(R) Xeon(R) Platinum 8280 CPUs, Nvidia GeForce RTX 3090 graphics cards, and 256\,GB (8\,$\times$\,32\,GB) DDR4 2666\,MHz ECC memory. All methods are evaluated on identical hardware.

\begin{figure*}
\begin{lstlisting}[language=Python, caption={Branching Rule Template.}, label={rule_template}]
import numpy as np
import math
def candidate_variable_score(variable_features) -> float: 
    """
    Assign a score to this candidate variable reflecting the potential dual bound improvement resulting from branching on it.


    Args:
    variable_features: Feature vector of the candidate variable, 72 dimensions.

    Return:
    A score indicating how much the dual bound would improve if this variable were chosen as the branching variable.
    """
    return 0.0

\end{lstlisting}
\end{figure*}

\subsection{Prompt design}
\begin{figure*}
    
    \centering
    \includegraphics[width=0.9\linewidth]{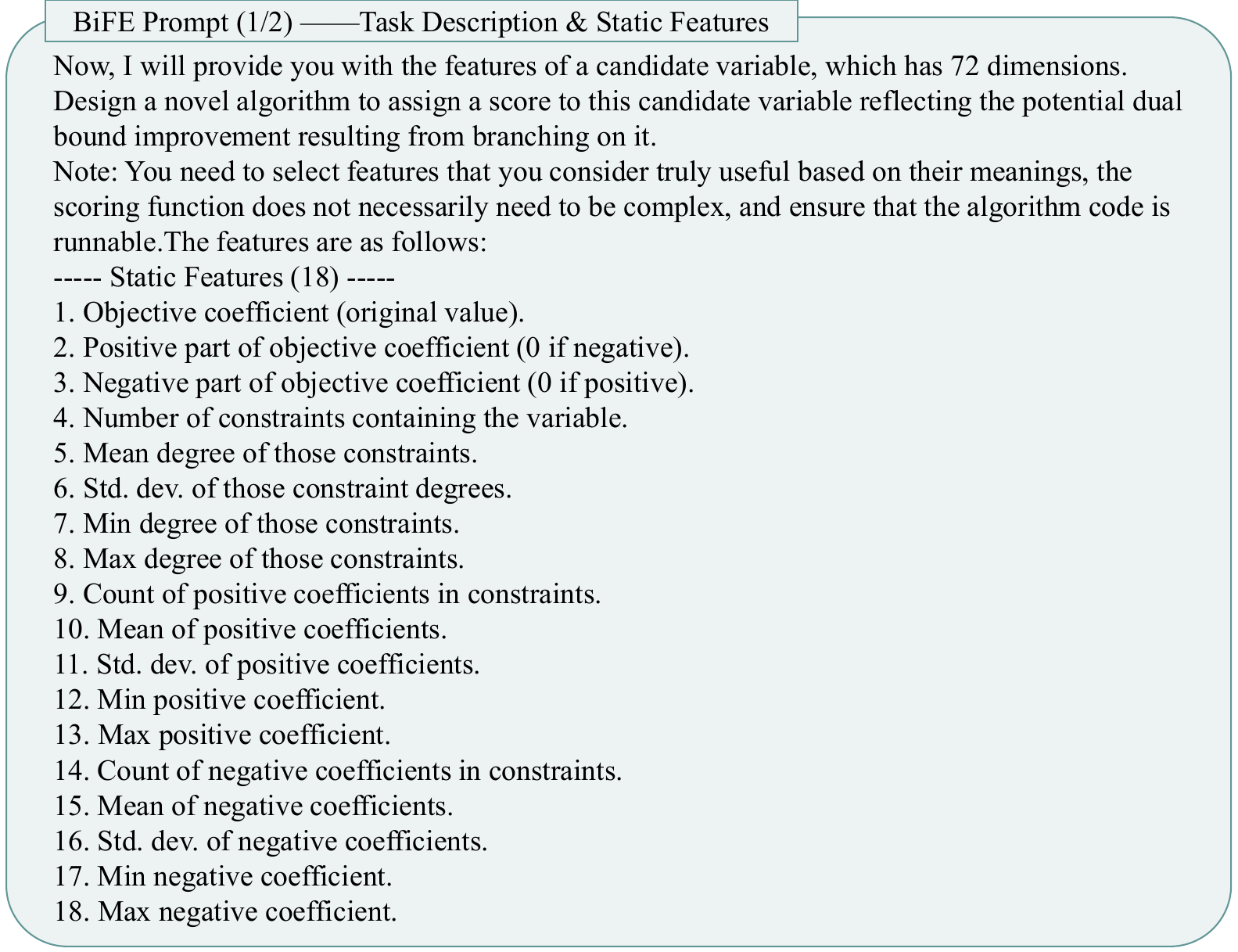}
    \caption{Problem Description Prompt for BiFE, Part I.}
    \label{fig:bife_prompt_1}
\end{figure*}

\begin{figure*}
    \centering
    \includegraphics[width=0.9\linewidth]{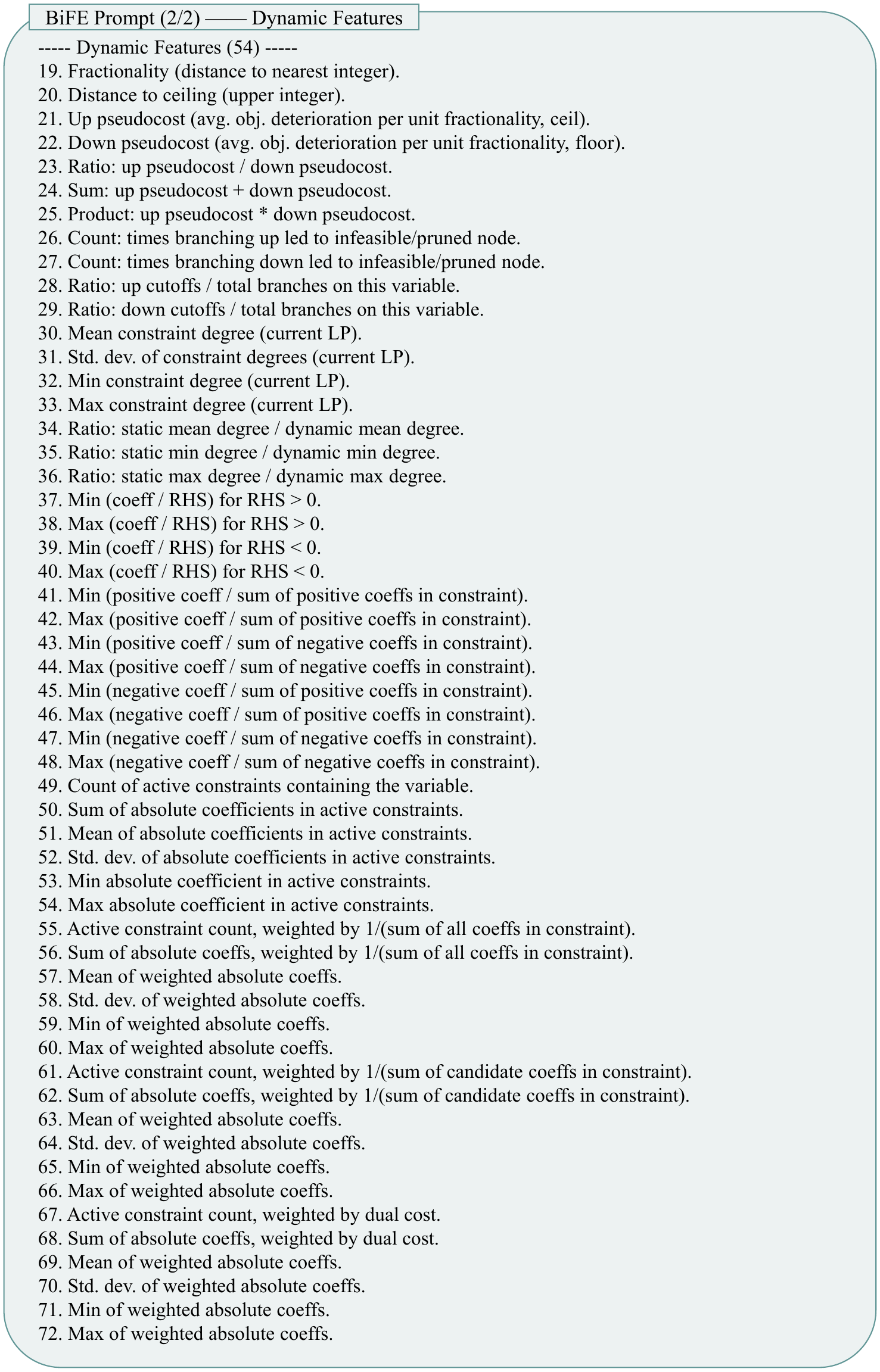}
    \caption{Problem Description Prompt for BiFE, Part II.}
    \label{fig:bife_prompt_2}
\end{figure*}

For LLM-based methods, the prompt templates for initialization, crossover, and mutation are identical to those used in EoH~\citep{liu2024EoH}, with one template for initialization, and two templates each for crossover and mutation.
The problem description prompts are illustrated in Figures~\ref{fig:bife_prompt_1} and~\ref{fig:bife_prompt_2}, and the rule template is shown in Listing~\ref{rule_template}. At each evolutionary operation, we concatenate the problem description, the corresponding operation prompt, and the parent selection results into a single input fed to the LLM, which then outputs a new heuristic for the subsequent generation.

\subsection{Evolutionary parameters}
The number of LLM calls, population size, and other evolutionary parameters are kept consistent across all LLM-based methods. The key hyperparameters are summarized in Table~\ref{tab:parameters}.

\begin{table}[t]
\caption{Hyperparameter settings of the evolutionary algorithm.}
\label{tab:parameters}
\centering
\begin{tabular}{lc}
\toprule
\textbf{Parameter} & \textbf{Value} \\
\midrule
Population size & 8 \\
Initial population size & 16 \\
Number of instances in $F_{HF}$ & 20 \\
Random seed per $F_{HF}$ instance & 0 \\
Maximum sampled heuristics (LLM calls) & 100 \\
LLM output time limit (s) & 60 \\
Number of parents in crossover & 2 \\
\bottomrule
\end{tabular}

\end{table}

\subsection{$\mathbf{F_{HF}}$ evaluation metrics}
For standard benchmarks (CA, SC, CFL, and MIS), we set a time limit of 60 seconds and use solving time as the primary performance metric, as these instances can typically be solved to optimality within this budget. For real-world benchmarks (NNV, OTS, and BIP), we impose a time limit of 300 seconds. Under this setting, all NNV instances can be solved to optimality, OTS instances can be partially solved, and BIP instances remain entirely unsolved. Consequently, we employ solving time, optimality gap, and PDI as the respective performance metrics for NNV, OTS, and BIP. The rationale is that for BIP instances, the optimality gap at 300 seconds may remain close to the initial large value, making gap comparisons less informative, whereas for OTS instances the gap has already been reduced to a relatively small value, allowing meaningful discrimination among methods. We therefore use gap for OTS and PDI for BIP to better capture algorithmic progress.

\subsection{LLM4Branch evaluation settings}
For a fair comparison, we set LLM4Branch's stage-1 screening, stage-2 validation, and Bayesian parameter-optimization limits to 60/60/10 seconds, respectively, for CA, SC, MIS, and CFL. For NNV, OTS, and BIP, the corresponding limits are 100/200/10 seconds. The Bayesian optimization uses two training instances per trial. We run 100 evolutionary iterations targeting solving time, with four training instances, 20 validation instances. Other search settings follow the released configuration: numerical parameter optimization is enabled; the population size, archive size, and number of islands are 120, 20, and 8; the elite-selection and exploitation ratios are 0.2 and 0.7. The dataset-specific Bayesian-optimization budgets are 80/10 calls/initial points for SC, 100/10 for CA, 50/10 for MIS, 16/2 for CFL, and 10/4 for each real-world benchmark.

\begin{table}[t]
\caption{Exact two-sided Wilcoxon signed-rank tests comparing instance-level solving time of BiFE and LLM4Branch. Each test uses 20 instance-level geometric means, obtained by aggregating five seeds per instance. Values are Holm-adjusted $p$-values across all eight comparisons; \textbf{bold} values are below 0.05.}
\label{tab:wilcoxon_bife_llm4branch}
\centering
\small
\setlength{\tabcolsep}{3mm}
\begin{tabular}{lcc}
\toprule
Setting & Holm-adjusted $p$ & BiFE win/tie/loss \\
\midrule
CA Test       & 0.4128 & 12/0/8 \\
SC Test       & 0.6146 & 15/0/5 \\
MIS Test      & 0.9354 & 12/0/8 \\
CFL Test      & \textbf{0.0018} & 18/0/2 \\
CA Transfer   & \textbf{0.0499} & 17/0/3 \\
SC Transfer   & 1.0000 & 10/0/10 \\
MIS Transfer  & 1.0000 & 10/0/10 \\
CFL Transfer  & \textbf{\textless
0.001} & 18/0/2 \\
\bottomrule
\end{tabular}

\end{table}

\section{Statistical Significance Tests}
\label{appendix:statistical_tests}

We compare BiFE with LLM4Branch using solving time, the primary metric for the standard benchmarks. For each benchmark setting, we aggregate the five seed-level solving times of each instance by their geometric mean, yielding 20 paired instance-level observations. This avoids treating repeated seeds of the same instance as independent samples. We then conduct exact two-sided Wilcoxon signed-rank tests. The displayed Holm-adjusted $p$-value controls the family-wise error rate over all eight comparisons. A value below 0.05 indicates statistically significant evidence that the paired solving times differ; the direction is reported through the win/tie/loss counts, where a win denotes a lower BiFE solving time.

Table~\ref{tab:wilcoxon_bife_llm4branch} shows that BiFE is significantly faster than LLM4Branch on CFL test instances, CA transfer instances, and CFL transfer instances in the statistical sense. In these settings, BiFE is faster on 17 or 18 of the 20 paired instances. The remaining five settings do not show a statistically significant difference after Holm correction. Nevertheless, BiFE records more instance-level wins than LLM4Branch in three of these settings (CA, SC, and MIS test instances), and ties LLM4Branch in the other two transfer settings; it is never outperformed in the win/tie/loss comparison. 
Importantly, our comparison is not intended to claim that BiFE uniformly and significantly improves the downstream solving performance of the discovered rules over LLM4Branch. Instead, the central comparison concerns the rule-design time required to obtain these rules: as shown in the main paper, BiFE requires substantially less search time than LLM4Branch while achieving comparable solving performance overall, with statistically significant improvements on several benchmarks. Thus, BiFE's primary advantage is its markedly higher search efficiency, which enables it to discover competitive branching rules---and, on some benchmarks, significantly superior ones---without incurring LLM4Branch's substantially higher rule-design cost.


\section{Rule Complexity Analysis and Examples}
\label{appendix:complexity}

\subsection{Motivation and overview}
\begin{figure}[t]
    \centering
    \includegraphics[width=\linewidth]{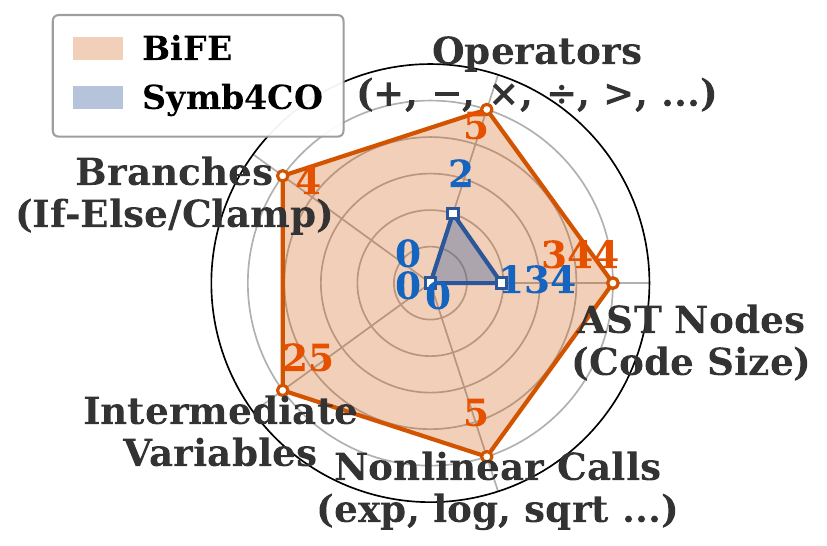}
    \caption{Complexity comparison between the branching rules discovered by Symb4CO and BiFE on SC.}
    \label{fig:appendix_rule_complexity}
\end{figure}

Across transfer and real-world benchmarks, symbolic rules exhibit a larger performance drop than BiFE in the main paper. To investigate a possible structural explanation, we compare the branching rules discovered by Symb4CO and BiFE on SC along five complementary complexity metrics. As Figure~\ref{fig:appendix_rule_complexity} shows, BiFE's rule is consistently more complex across these metrics, indicating a greater capacity to express multi-step algorithmic logic. The following definitions and rule examples make this comparison explicit.

\subsection{Metric definitions}
To quantitatively compare the complexity of branching rules discovered by different methods, we define five metrics:
\begin{itemize}
\item \textbf{AST Node Count:} The total number of nodes in the Abstract Syntax Tree of the generated code, capturing overall code size and structural complexity.
\item \textbf{Operator Count:} The number of distinct arithmetic and comparison operators (e.g., \texttt{+}, \texttt{-}, \texttt{*}, \texttt{/}, \texttt{<=}, \texttt{>}) appearing in the rule, reflecting the diversity of mathematical operations employed.
\item \textbf{Non-linear Calls:} The number of invocations of non-linear functions such as \texttt{exp}, \texttt{log}, \texttt{log1p}, \texttt{sqrt}, and \texttt{abs}, indicating the rule's ability to capture non-linear relationships.
\item \textbf{Branches:} The total count of if-else statements and clamp operations (e.g., \texttt{max}, \texttt{min}), quantifying the rule's conditional logic and decision-making capacity.
\item \textbf{Intermediate Variables:} The number of named variables that store intermediate computation results, reflecting the stepwise reasoning structure embedded in the rule.
\end{itemize}


\subsection{Illustrative examples}
On standard benchmarks, the rules discovered by BiFE and Symb4CO exhibit clear differences in complexity.
The rules found by BiFE are shown in Listing~\ref{rule_CA} (CA), Listing~\ref{rule_SC} (SC), Listing~\ref{rule_MIS} (MIS) and Listing~\ref{rule_CFL} (CFL).
For comparison, the rules discovered by Symb4CO on the same benchmarks are provided in
Listing~\ref{rule_symb4co_CA}, Listing~\ref{rule_symb4co_SC}, Listing~\ref{rule_symb4co_MIS}, Listing~\ref{rule_symb4co_CFL}.
Even from a purely visual inspection, BiFE's rules exhibit markedly more sophisticated structure than those of Symb4CO. Taking the rules discovered on CA as an example, BiFE's rule (Listing~\ref{rule_CA}) first extracts and caps several features, computes a harmonic mean of pseudocosts, combines them with a weighted sum via a geometric mean, and finally applies a logistic transformation to spread the scores---demonstrating a clear, multi-step algorithmic logic.
In contrast, Symb4CO's rule (Listing~\ref{rule_symb4co_CA}) simply multiplies a sequence of input features together in a single expression, with no intermediate reasoning or non-linear processing.
This qualitative contrast aligns with the quantitative comparison in Figure~\ref{fig:appendix_rule_complexity}, suggesting that LLM-based methods inherently generate more complex algorithmic logic suited for challenging MILPs.

\begin{figure*}[t]
\begin{lstlisting}[language=Python, caption={Branching Rule Designed by BiFE on CA.}, label={rule_CA}]
import numpy as np
import math
def candidate_variable_score(variable_features) -> float:
    """
    Assign a score to this candidate variable reflecting the potential dual bound improvement resulting from branching on it.


    Args:
    variable_features: Feature vector of the candidate variable, 72 dimensions.

    Return:
    A score indicating how much the dual bound would improve if this variable were chosen as the branching variable.
    """
    f = variable_features
    frac = max(f[18], 1e-10)
    up_pc = max(f[20], 1e-10)
    down_pc = max(f[21], 1e-10)
    harmonic_pc = 2.0 / (1.0/up_pc + 1.0/down_pc)
    weighted_active_sum = max(f[67], 1e-10)
    
    geom_part = (frac * harmonic_pc) ** 0.5
    logit_input = geom_part * weighted_active_sum
    # Inverse logistic to center and spread scores
    raw = logit_input / (logit_input + 1.0)
    return 1.0 / (1.0 + np.exp(-5.0 * (raw - 0.5)))
\end{lstlisting}
\end{figure*}

\begin{figure*}[t]
\begin{lstlisting}[language=Python, caption={Branching Rule Designed by BiFE on SC.}, label={rule_SC}]
import numpy as np
import math
def candidate_variable_score(variable_features) -> float:
    """
    Assign a score to this candidate variable reflecting the potential dual bound improvement resulting from branching on it.


    Args:
    variable_features: Feature vector of the candidate variable, 72 dimensions.

    Return:
    A score indicating how much the dual bound would improve if this variable were chosen as the branching variable.
    """
    import math
    
    fractionality = variable_features[18]  # feature 19
    up_pseudocost = variable_features[20]  # feature 21
    down_pseudocost = variable_features[21]  # feature 22
    weighted_active_count = variable_features[66]  # feature 67
    unweighted_active_count = variable_features[48]  # feature 49
    
    # Geometric mean of pseudocosts
    if up_pseudocost <= 0 or down_pseudocost <= 0:
        pseudocost_geom = 0.0
    else:
        pseudocost_geom = math.sqrt(up_pseudocost * down_pseudocost)
    
    # Weighted harmonic mean of active constraint counts (avoid division by zero)
    if weighted_active_count > 0 and unweighted_active_count > 0:
        harmonic_mean = 2.0 / (1.0 / weighted_active_count + 1.0 / unweighted_active_count)
    else:
        harmonic_mean = max(weighted_active_count, unweighted_active_count)
    
    # Product with fractionality, then log-scale clamp
    raw_score = fractionality * pseudocost_geom * (1.0 + harmonic_mean)
    
    # Normalize via log transformation and clamp to [0, 1]
    if raw_score <= 0:
        return 0.0
    normalized = math.log(1.0 + raw_score) / math.log(1.0 + 100.0)
    return min(1.0, normalized)
\end{lstlisting}
\end{figure*}

\begin{figure*}[t]
\begin{lstlisting}[language=Python, caption={Branching Rule Designed by BiFE on MIS.}, label={rule_MIS}]
import numpy as np
import math
def candidate_variable_score(variable_features) -> float:
    """
    Assign a score to this candidate variable reflecting the potential dual bound improvement resulting from branching on it.


    Args:
    variable_features: Feature vector of the candidate variable, 72 dimensions.

    Return:
    A score indicating how much the dual bound would improve if this variable were chosen as the branching variable.
    """
    frac = variable_features[18]
    up_pseudo = variable_features[20]
    down_pseudo = variable_features[21]
    pseudo_sum = variable_features[23]
    active_count = variable_features[48]
    sum_abs_dual = variable_features[67]

    # Fractionality component: tanh scaling
    frac_score = math.tanh(frac * 3.0)

    # Pseudocost component: geometric mean normalized by (1 + sum)^0.5
    if up_pseudo > 1e-9 and down_pseudo > 1e-9:
        geo_mean = math.sqrt(up_pseudo * down_pseudo)
    else:
        geo_mean = max(up_pseudo, down_pseudo)
    pseudo_score = geo_mean / (1.0 + math.sqrt(pseudo_sum))

    # Activity component: product of active count and dual-weighted sum, log-scaled
    if active_count > 0 and sum_abs_dual > 0:
        activity_score = math.log1p(active_count * sum_abs_dual)
    else:
        activity_score = 0.0

    # Weighted sum and compress with square root
    score = (2.5 * frac_score + 3.5 * pseudo_score + 2.0 * activity_score)
    return math.sqrt(max(score, 0.0))
\end{lstlisting}
\end{figure*}

\begin{figure*}[t]
\begin{lstlisting}[language=Python, caption={Branching Rule Designed by BiFE on CFL.}, label={rule_CFL}]
import numpy as np
import math
def candidate_variable_score(variable_features) -> float:
    """
    Assign a score to this candidate variable reflecting the potential dual bound improvement resulting from branching on it.


    Args:
    variable_features: Feature vector of the candidate variable, 72 dimensions.

    Return:
    A score indicating how much the dual bound would improve if this variable were chosen as the branching variable.
    """
    fractionality = variable_features[18]          # Feature 19
    up_pc = variable_features[20]                 # Feature 21
    down_pc = variable_features[21]               # Feature 22
    count_pos = variable_features[8]              # Feature 9: count of positive coefficients in constraints
    count_neg = variable_features[13]             # Feature 14: count of negative coefficients in constraints
    
    # Avoid division by zero and extreme values
    pos = max(count_pos, 1e-8)
    neg = max(count_neg, 1e-8)
    
    # Balance ratio: closer to 1 means balanced positive/negative presence
    ratio = min(pos, neg) / max(pos, neg)
    
    # Pseudocost product (feature 25)
    pc_product = max(up_pc * down_pc, 1e-8)
    
    # Score: fractionality * sqrt(pseudocost product) * balance ratio
    score = fractionality * (pc_product ** 0.5) * ratio
    
    return max(score, 0.0)
\end{lstlisting}
\end{figure*}

\begin{figure*}[t]
\begin{lstlisting}[language=Python, caption={Branching Rule Designed by Symb4CO on CA.}, label={rule_symb4co_CA}]
def candidate_variable_score(input)
    return ((((((((((((inputs[:,89] * inputs[:,89]) * inputs[:,84]) * inputs[:,35]) * inputs[:,89]) * inputs[:,89]) * inputs[:,67]) * inputs[:,21]) * inputs[:,16]) * inputs[:,57]) * inputs[:,89]) * inputs[:,89]) * inputs[:,84])
\end{lstlisting}
\end{figure*}

\begin{figure*}[t]
\begin{lstlisting}[language=Python, caption={Branching Rule Designed by Symb4CO on SC.}, label={rule_symb4co_SC}]
def candidate_variable_score(input)
    return (((((((((((inputs[:,79] * inputs[:,26]) + inputs[:,84]) + inputs[:,57]) + inputs[:,70]) + inputs[:,70]) + inputs[:,70]) * inputs[:,79]) + inputs[:,37]) * inputs[:,37]) * inputs[:,89]) * inputs[:,89])
\end{lstlisting}
\end{figure*}

\begin{figure*}[t]
\begin{lstlisting}[language=Python, caption={Branching Rule Designed by Symb4CO on MIS.}, label={rule_symb4co_MIS}]
import torch
def candidate_variable_score(input)
    return ((((((((((((inputs[:,89] + inputs[:,32]) + inputs[:,71]) * inputs[:,90]) * inputs[:,32]) + inputs[:,90]) + inputs[:,89]) + inputs[:,90]) + inputs[:,32]) + inputs[:,89]) - inputs[:,22]) + torch.tensor(0.2, dtype=torch.float, device=consts.DEVICE)) + inputs[:,75])
\end{lstlisting}
\end{figure*}

\begin{figure*}[t]
\begin{lstlisting}[language=Python, caption={Branching Rule Designed by Symb4CO on CFL.}, label={rule_symb4co_CFL}]
def candidate_variable_score(input)
    return ((((((((((inputs[:,71] * inputs[:,71]) + inputs[:,24]) * inputs[:,34]) + inputs[:,19]) + inputs[:,71]) + inputs[:,71]) + inputs[:,71]) + inputs[:,24]) + inputs[:,89]) * inputs[:,89])
\end{lstlisting}
\end{figure*}

\end{document}